# BSC-Net: A Small-Branch-Sensitive Structural Continuity Network for Coronary Vessel Segmentation and Quantitative Angiographic Analysis

Wanxian Li[1,#], Jiaqian Qin[1,#], Qingyi Xian[1], Yazhi Li[1], Song Chen[2], Liman Li[3], Hao He[1*]

[1]School of Biomedical Engineering, Sun Yat-sen University, Shenzhen, 518107, China.

[2]Department of Cardiovascular Surgery, Zhongnan Hospital of Wuhan University, Wuhan, 430071, China

[3]Department of Laboratory Medicine, West China Hospital of Sichuan University, Chengdu, 610041, China.

[†] Correspondence: H.H. (hehao23@mail.sysu.edu.cn)

# These authors contributed equally.

## Abstract

Vessel segmentation in X-ray coronary angiography (XCA) is a fundamental step for quantitative coronary analysis and subsequent assessment of coronary artery disease. However, accurate vessel segmentation remains challenging because of imaging noise, complex bifurcations, and the overlap of vessels and background structures, which can lead to disrupted vascular connectivity and missed small branches. In this work, we propose BSC-Net, a ResNet–U-Net-based framework tailored to improve small-vessel representation and repair vascular structural continuity. BSC-Net enhances small-vessel representation through targeted sampling and improves vascular structural continuity by integrating long-range contextual modeling and Edge-Informed Loss (EIL). BSC-Net was validated on two public XCA datasets, demonstrating state-of-the-art (SOTA) performance in coronary vessel segmentation with Dice and IoU scores of 77.8%/90.6% and 64.5%/83.0%, respectively. Furthermore, based on the obtained vessel segmentation, we performed automated quantitative coronary analysis and derived clinically relevant morphological and hemodynamic parameters, including stenosis ratio, time-to-peak, and relative propagation velocity. These results demonstrate that BSC-Net produces accurate vessel segmentation results with preserved vascular continuity for quantitative coronary assessment, enabling reliable downstream analysis and clinical evaluation of coronary artery disease.

## 1. Introduction

Accurate coronary vessel segmentation in X-ray coronary angiography (XCA) is a fundamental prerequisite for automated quantitative coronary analysis and computer-assisted assessment of coronary artery disease. Accurate delineation of vessel boundaries and centerlines enables quantitative measurements of vessel diameter and stenosis severity and provides the spatial basis for analyzing contrast-agent propagation during coronary intervention [1], [2]. The reliability of these quantitative analyses depends largely on the accuracy of the underlying vascular segmentation. Inaccurate boundary localization or locally interrupted vascular structures can propagate errors into subsequent morphometric measurements and contrast-propagation analysis. Therefore, XCA segmentation requires not only high overlap with the ground-truth vessel mask but also accurate vessel boundaries and continuous vascular structures [3].

Traditional coronary vessel extraction methods can be broadly divided into image-processing approaches and machine-learning-based approaches [4], [5]. Image-processing methods identify vascular structures according to predefined intensity, geometric, or morphological characteristics, including thresholding, region growing, morphological processing, *Frangi* filtering based on Hessian eigenvalue analysis [6], and directional tracking according to local orientation continuity [7]. Conventional machine-learning methods typically formulate vessel extraction as a pixel- or patch-level classification problem, combining classifiers such as support vector machines, random forests, or boosting models with handcrafted intensity, gradient, texture, and structural features [8]–[10]. Although these approaches can effectively identify vessels with relatively high contrast and clear boundaries, their reliance on predefined rules or handcrafted local features limits their ability to distinguish weakly contrasted vessels from complex anatomical backgrounds, particularly in distal coronary branches.

Deep learning has substantially improved XCA vessel segmentation by learning hierarchical vascular representations directly from images. U-Net and its variants employ multiscale encoder–decoder architectures with skip connections to integrate high-level semantic information with fine spatial details, enabling more accurate vessel delineation than conventional feature-based approaches [11]–[13]. XCA-specific convolutional networks, such as AngioNet and related models, further adapt deep segmentation architectures to coronary angiography by incorporating learnable image-processing or feature-enhancement components, thereby improving the discrimination of vascular structures from complex angiographic backgrounds [14], [15]. Despite these advances, CNN-based XCA segmentation still faces several limitations. First, large and proximal vessels are generally easier to segment because they occupy more pixels and exhibit stronger contrast and clearer boundaries, whereas weakly contrasted distal branches contribute less effectively to feature learning and are therefore more prone to be missed. Second, conventional CNNs mainly capture local contextual information through convolutional operations, limiting their ability to model long-range dependencies across the vascular tree. Third, commonly used pixel-wise or overlap-based loss functions mainly emphasize regional segmentation accuracy, providing limited constraints on vascular continuity and local structural consistency. Consequently, fine peripheral branches may remain incomplete, with local discontinuities and structural distortions in the predicted vascular tree.

Recent studies have attempted to address these limitations from several directions. For weakly contrasted small vessels, coarse-to-fine and multiscale strategies have been explored to enhance the representation of fine vascular structures. For example, Thuy et al. adopted a coarse-to-fine framework that further refines small-vessel regions after major-vessel segmentation, while Jiang et al. introduced multiresolution inputs and multiscale convolution to improve the representation of vessels with varying calibers and contrast [16], [17]. Although these strategies improve the sensitivity to small vascular structures, they usually require additional refinement stages or multiscale processing, increasing the complexity of the segmentation pipeline. Beyond enhancing vessel representation, preserving the overall organization of the vascular tree also requires broader contextual information. Multiscale context networks such as CE-Net and CPFNet enlarge the effective receptive field and aggregate contextual information through dense atrous convolution, multi-kernel pooling, and hierarchical feature fusion [18]–[20]. These designs enrich multiscale contextual representations and improve the delineation of vascular structures across different spatial scales. However, because feature interactions are still mainly established through convolutional and multiscale aggregation operations, their ability to explicitly model dependencies between distant vascular regions remains limited. Transformer-based architectures, including TransUNet, TransFuse, and Swin-Unet, further introduce self-attention mechanisms to capture global or long-range contextual relationships and enhance structural modeling [21]–[24]. However, applying attention to high-resolution feature maps can substantially increase computational complexity. In addition to improving feature representation, vascular structural preservation has also been addressed through the design of loss functions. Centerline-, boundary-, topology-, and connectivity-aware loss functions, including clDice and skeleton-recall-based formulations, provide additional structural supervision for thin vascular structures and improve the preservation of vascular centerlines and topology [25]–[32]. Nevertheless, these loss functions provide limited joint constraints on local branch continuity, boundary alignment, and vessel-caliber consistency. Overall, existing approaches have improved small-vessel representation, contextual modeling, and vascular structural preservation from different perspectives. However, small-vessel enhancement often introduces additional processing, long-range contextual modeling remains computationally demanding, and local anatomical consistency remains insufficiently constrained.

To address these challenges, we propose **BSC-Net**, a hybrid segmentation framework that integrates ResNet and U-Net to achieve accurate coronary artery segmentation with improved structural continuity. To capture long-range vascular dependencies with limited computational overhead, **BSC-Net** introduces a Swin Transformer at the low-resolution bottleneck. In addition, we design an **Edge-Informed Loss (EIL)** to improve vessel-boundary alignment and enhance the structural continuity of fine vascular branches. Furthermore, a small-vessel-aware sampling strategy is employed to increase the representation of small and weakly contrasted vessels during training, thereby encouraging the model to pay greater attention to fine vascular structures. Benefiting from these optimizations, **BSC-Net** improves long-range vascular modeling, structural continuity and strengthens the learning of fine vascular structures. The overall framework is illustrated in Fig. 1, while implementation details of the sampling strategy are provided in the Supplementary Information.

Experiments on the public MOSXAV XCA benchmark demonstrate that BSC-Net achieves state-of-the-art segmentation performance, with a Dice score of **77.80%** and an IoU of **64.53%**, outperforming Swin-UMamba by **3.60** and **4.40 percentage points**, respectively. Structural evaluation further shows reductions of **13.7%** in connected-component count error and **15.7%** in endpoint count error, indicating improved preservation of vessel-tree continuity. Moreover, the improved segmentation results facilitate the quantitative estimation of vascular anatomical and hemodynamic-related parameters, thereby supporting computer-assisted clinical diagnosis and interventional evaluation of coronary artery disease.

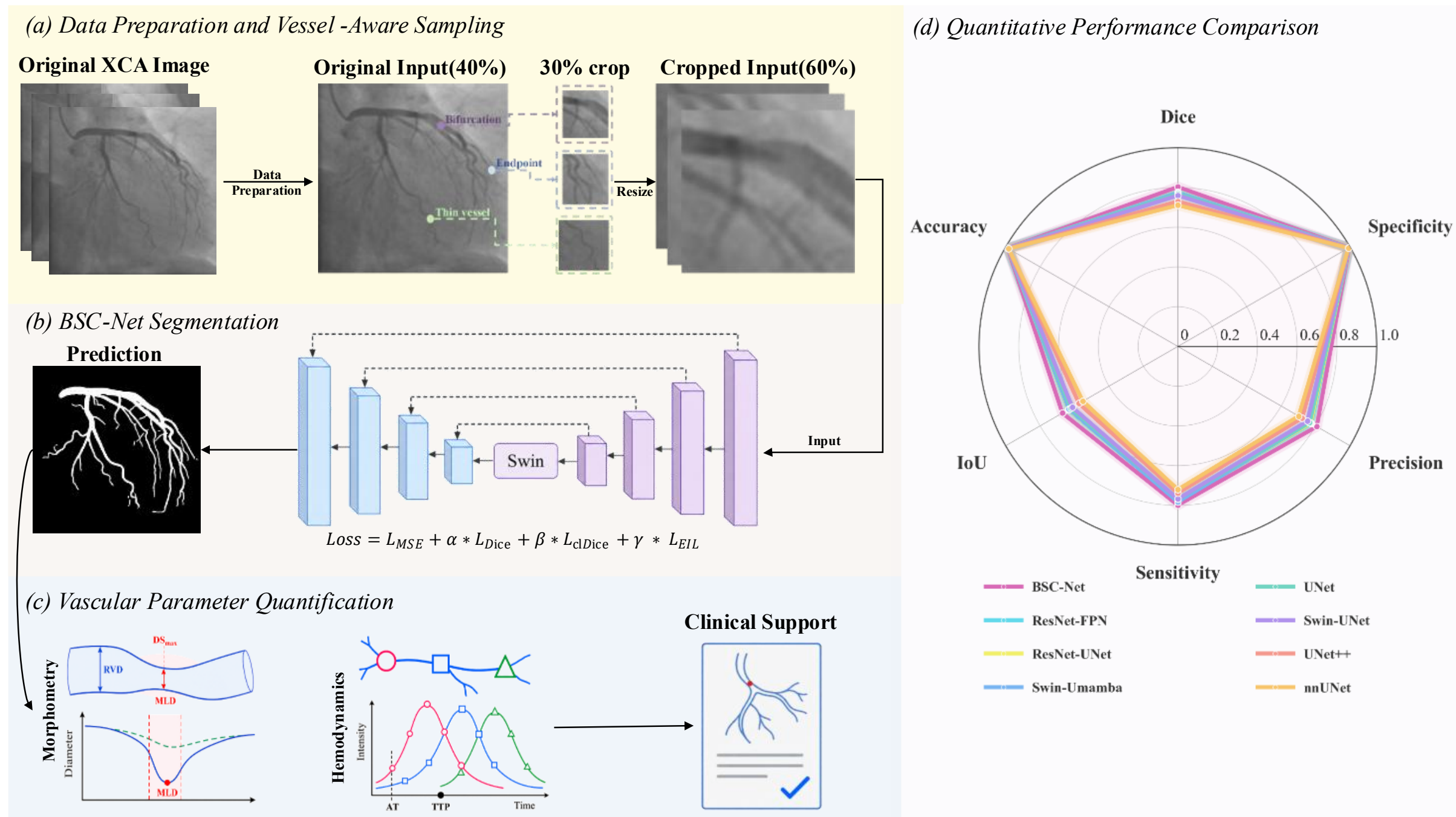

**Fig. 1 | Workflow of the proposed BSC-Net framework, downstream vascular analysis, and quantitative performance comparison**. (**a**) Data preparation and vessel-aware sampling for constructing training samples. (**b**) Coronary artery segmentation using BSC-Net. (**c**) Vascular parameter quantification, including morphological and hemodynamic analysis. (**d**) Multi-metric comparison of BSC-Net and representative segmentation models in terms of Dice, accuracy, specificity, precision, sensitivity, and IoU.

## 2. BSC-Net Method

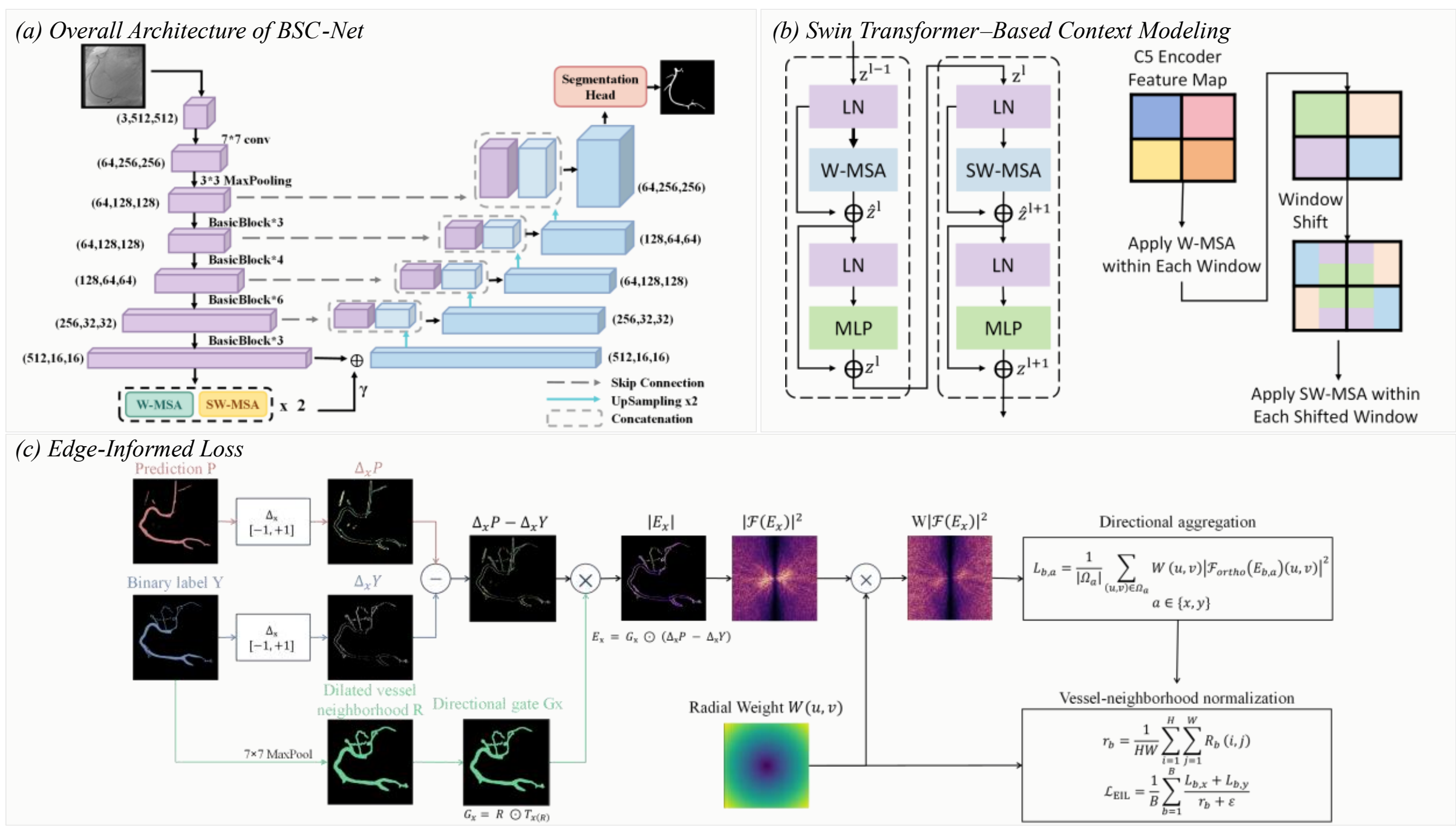

**Fig.2 | Network structure of BSC-Net.** (**a**) Overall encoder–decoder architecture of BSC-Net with a Swin Transformer bottleneck. (**b**) Swin Transformer–based context modeling through W-MSA and SW-MSA. (**c**) Edge-informed loss (EIL) for enhancing edge consistency and vascular structural continuity.

As shown in Fig. 2, BSC-Net uses a ResNet-34 U-Net as the backbone, incorporates small-vessel-aware sampling during training, introduces EIL at the loss level, and places a Swin Transformer only at the low-resolution bottleneck. These three components target increased exposure to thin and structurally vulnerable vessel regions, local structural consistency, and broader contextual modeling, respectively.

## 2.1 Small-Vessel-Aware Sampling Strategy

To increase the representation of small and structurally vulnerable vessel regions during training, we use endpoints, bifurcation points, and thin-vessel regions as candidates for local cropping. Endpoints and bifurcations are identified from the 8-neighborhood degree of the skeleton, whereas thin-vessel points are determined from the lower-quantile region of the Euclidean distance transform (EDT) on the skeleton. This increases the occurrence frequency of easily missed structures in the training samples.

Given a binary vessel label of size $H \times W$, denoted $Y$, morphological skeletonization is first applied to obtain the vessel skeleton $S = \mathrm{Skel}(Y)$. For any skeleton point $p \in S$, let $\mathcal{N}_8(p)$ denote its 8-neighborhood, and compute the local degree using $d(p) = |\mathcal{N}_8(p) \cap S|$. Skeleton points with degree 1 and degree no less than 3 form the endpoint set and bifurcation set, respectively. The Euclidean distance transform is then used to compute the distance $R(p) = \mathrm{DT}(Y)(p)$ from each skeleton point to the nearest background pixel, and the 25th percentile $r_{25} = Q_{0.25}(\{R(p) \mid p \in S\})$ of the distance values over all skeleton points is used as the threshold for identifying thin-vessel points. The three candidate sets are defined as:

$$\begin{aligned} P_{\mathrm{end}} &= \{p \in S \mid d(p) = 1\}, \\ P_{\mathrm{branch}} &= \{p \in S \mid d(p) \geq 3\}, \\ P_{\mathrm{thin}} &= \{p \in S \mid R(p) \leq r_{25}\} \end{aligned} \tag{1}$$

where $P_{\mathrm{end}}$, $P_{\mathrm{branch}}$, and $P_{\mathrm{thin}}$ denote the endpoint, bifurcation-point, and thin-vessel-point sets, respectively. The final sampling candidate set is:

$$P_{\mathrm{cand}} = P_{\mathrm{end}} \cup P_{\mathrm{branch}} \cup P_{\mathrm{thin}}. \tag{2}$$

During training, full-image and vessel-centered local-crop samples are used in complementary proportions. A full image is selected with probability $p_{full}$ = 0.40, whereas local cropping is selected with probability $p_{crop}$ = 0.60. The local crop scale is set to 30% of the original image extent and is resized to the unified 512×512 input size. Full images preserve learning of the overall vessel tree, whereas local crops strengthen supervision of endpoints, bifurcations, and thin-vessel regions. Because both modes are ultimately presented to the network at the same input size, the strategy does not change the number of network parameters or introduce additional inference steps.

## 2.2 BSC-Net Architecture

As shown in Fig. 2(a), BSC-Net employs an ImageNet-pretrained ResNet-34 encoder and a four-stage U-shaped decoder. The encoder extracts multiscale features, while the decoder progressively upsamples them and fuses the corresponding skip features. A segmentation head finally outputs the vessel probability map.

The deepest bottleneck feature has a spatial size of 16×16 and 512 channels. A Swin Transformer module is inserted only at this scale and fused with the original CNN bottleneck feature through a learnable residual connection before decoding. The detailed configuration is provided in Supplementary Material S6.

## 2.3 Swin Transformer Context Modeling

As shown in Fig. 2(b), BSC-Net places a Swin Transformer [33] at the low-resolution 16×16 bottleneck to capture broader vascular context while limiting attention computation. The module alternates window-based multi-head self-attention (W-MSA) and shifted-window multi-head self-attention (SW-MSA).
The bottleneck feature is projected from 512 channels to 384 dimensions using a 1×1 convolution and represented as 256 tokens on a 16×16 grid. A window size of M = 4 partitions this grid into 16 non-overlapping 4×4 windows.

In each W-MSA block, self-attention is computed independently within each window. After W-MSA, each SW-MSA block enables interactions across window boundaries by cyclically shifting the feature grid by half a window, corresponding to two token positions in both the horizontal and vertical directions. The shifted grid is repartitioned into 4×4 windows for masked self-attention, allowing tokens from previously separate windows to interact. The cyclic shift is then reversed to restore the original spatial arrangement.

Four Swin blocks are stacked in an alternating W-MSA/SW-MSA sequence. The final token sequence is reshaped into a two-dimensional feature map and projected back to 512 channels using a 1×1 convolution. This feature map is fused with the original CNN bottleneck feature through a learnable residual connection:

$$\Delta = \mathrm{Conv}_{1\times1}\left[\mathrm{Reshape}\big(\mathrm{Swin}(T_0)\big)\right], \qquad C'_5 = C_5 + \gamma_s \Delta, \qquad \gamma_s^{(0)} = 0.1. \tag{3}$$

where γ is a learnable residual scaling coefficient initialized to 0.1. Repeated within-window attention and shifted-window interactions propagate contextual information beyond individual windows. Because bottleneck tokens correspond to relatively large input-image regions, this mechanism supports modeling of long-range vessel trajectories. Residual fusion incorporates this context while retaining the original CNN features. The attention and masking formulations follow the original Swin Transformer and are detailed in Supplementary Material S6.

2.4 Edge-Informed Loss

As shown in Fig. 2(c), local discontinuities, missing small branches, and boundary displacement typically involve only a small number of pixels and can therefore be underweighted by region-based losses. Motivated by the radial frequency weighting used in the Fourier-domain formulation of Elastic Interaction-Based Loss [34], we propose EIL to characterize local structural discrepancies through direction-aware gradient residuals within vessel neighborhoods and frequency-dependent weighting.
To jointly constrain pixel-wise prediction, region overlap, centerline coverage, and local gradient structure, BSC-Net combines EIL with MSE, Dice, and clDice:

$$\mathcal{L} = \mathcal{L}_{\mathrm{MSE}} + \alpha\mathcal{L}_{\mathrm{Dice}} + \beta\mathcal{L}_{\mathrm{clDice}} + \gamma\mathcal{L}_{\mathrm{EIL}} \tag{4}$$

Here, MSE provides pixel-wise supervision, Dice constrains region overlap, clDice constrains centerline coverage, and EIL supplements local structural supervision; the weights of the individual terms are determined on the validation set.

EIL first restricts structural comparisons to regions surrounding the annotated vessels. Specifically, a 7×7 max-pooling operation is applied to the ground-truth label to construct a vessel-neighborhood mask, thereby reducing the contribution of distant background regions to the structural term.

Within this neighborhood, one-pixel shifts are applied along the horizontal and vertical directions to construct directional gates. The corresponding gradient residuals between the predicted probability map P and label Y are then computed as:

$$G_a = R \odot \mathcal{T}_a(R), \qquad E_a = G_a \odot (\nabla_a P - \nabla_a Y), \qquad a \in \{x, y\} \tag{5}$$

The directional gates require both adjacent pixels involved in each gradient comparison to lie within the vessel neighborhood. Consequently, gradient discrepancies from distant background regions are suppressed, while local variations associated with vessel boundaries, branch endpoints, and discontinuities remain active. Boundary misalignment, missing branches, or local breaks therefore produce inconsistent directional responses between the prediction and ground truth, resulting in nonzero residuals in $E_x$ and $E_y$.

The directional residual maps are subsequently transformed using a normalized two-dimensional real fast Fourier transform. A radial weight is constructed according to the distance of each frequency location from the zero-frequency component:

$$\hat{E}_{a,b}(u, v) = \mathrm{RFFT2}_{\mathrm{ortho}}\big(E_{a,b}\big)(u, v), \qquad W(u, v) = \left[f_x(v)^2 + f_y(u)^2 + \varepsilon_f\right]^{\kappa/2} \tag{6}$$

The frequency-weighted gradient residual energy constitutes the core of EIL. For the sample(b), the weighted residual energy in direction (a) is defined as:

$$\mathcal{E}_{a,b} = \frac{1}{|\Omega_f|} \sum_{(u,v)\in\Omega_f} W\,(u, v)\left|\hat{E}_{a,b}(u, v)\right|^2 \tag{7}$$

Here, $\Omega_f$ denotes the set of discrete frequency locations in the two-dimensional real spectrum, and $|\Omega_f|$ is the number of frequency locations in this set. Spatially localized structural discrepancies tend to introduce stronger fine-scale components into the gradient-residual spectrum. Because $W(u, v)$ increases with radial frequency, these components receive greater relative weight, allowing local structural errors that occupy only a few pixels to contribute more effectively to the optimization objective.

Finally, the frequency-domain energies from the two directions are summed and normalized by the vessel-neighborhood occupancy ratio:

$$q_b = \frac{1}{HW} \sum_{i=1}^{H} \sum_{j=1}^{W} R_{b,i,j}\,, \qquad \mathcal{L}_{\mathrm{EIL}} = \frac{1}{B} \sum_{b=1}^{B} \frac{\mathcal{E}_{x,b} + \mathcal{E}_{y,b}}{\max(q_b, \varepsilon)} \tag{8}$$

This normalization reduces variation in the magnitude of the structural term caused by differences in vessel-neighborhood occupancy across samples. Overall, EIL combines vessel-neighborhood restriction, direction-aware gradient comparison, and frequency-dependent weighting to improve sensitivity to local structural discrepancies that may be insufficiently represented by region-overlap losses. Complete computation and implementation details are provided in Supplementary Material S5.

## 3. Experiments and Results

### 3.1 Datasets, Implementation Details, and Evaluation Metrics

Experiments were conducted on two public XCA datasets, MOSXAV and ICA_NJ. MOSXAV contains 30 dynamic XCA sequences with 1,789 frames, including 844 paired image/mask annotations. Device-related annotation discontinuities were automatically corrected before sample construction. Peak-opacification frames were identified according to vessel foreground area, and five-frame windows centered on these frames were extracted, yielding 222 static image–mask pairs (146 training and 76 test samples). ICA_NJ contains 616 paired static images; following the original 4:1 partition, 492 images formed the training pool and 124 the test set, with the training pool further divided into 393 training and 99 validation images. All samples followed the original sequence/case-level split protocol and were resized to $512 \times 512$, with geometric and intensity augmentation applied during training. Further preprocessing details are provided in Supplementary Material S1.

The model was implemented in PyTorch 2.6.0 and trained for 50 epochs on a single RTX 3090 using AdamW with a batch size of 2. The initial learning rate and weight decay were both $1 \times 10^{-4}$, with cosine annealing to $1 \times 10^{-6}$. The ImageNet-pretrained ResNet-34 encoder was fine-tuned end-to-end. The Swin residual coefficient $\gamma_s$ was initialized to 0.1, and the loss weights for Dice, clDice, and EIL were set to 0.35, 0.16, and 0.05, respectively. The post-processing threshold $\tau_s$ was set to 0.20 and was used only for downstream quantitative analysis, not for the segmentation results in Tables 1–3.

Comparison methods included U-Net [11], UNet++ [12], nnU-Net [35], ResUNet, ResNet-FPN, Swin-UNet [24], and Swin-UMamba [36]. ResUNet and ResNet-FPN were implemented in-house using standard architectures. Evaluation included Dice, IoU, Sensitivity, Precision, clDice, main-trunk recall, and fine-branch recall. Additional topology-related metrics are described in Supplementary Materials S1 and S3.

### 3.2 Comparison with Baseline Methods

As shown in Table 1, BSC-Net achieved an IoU of 64.53% and a Dice score of 77.80% on the more challenging MOSXAV dataset, outperforming the strongest baseline by 4.40 and 3.60 percentage points, respectively. Precision reached 76.72%, indicating that the improvement was not obtained simply by expanding the predicted foreground region.

On ICA_NJ, BSC-Net achieved the best Dice score of 90.60% and the best IoU of 83.00%, exceeding the strongest competing results by 0.50 and 0.70 percentage points, respectively. Sensitivity reached 90.50%, tying Swin-UMamba for the highest value, while Precision reached 90.90%, only 0.10 percentage points below the best result. These results indicate that BSC-Net remains competitive across multiple metrics even on this high-performance benchmark.

Fig. 3 further shows that BSC-Net provides more complete vessel coverage and fewer fragmented false detections around weakly opacified small branches and complex bifurcations on MOSXAV, while maintaining stable structural recovery on ICA_NJ. On MOSXAV samples with weakly opacified small branches and complex backgrounds, BSC-Net provides more complete vessel coverage and fewer scattered false detections. On ICA_NJ, where overall segmentation performance is already high, differences among methods become narrower, while BSC-Net still maintains stable vessel-structure recovery. These enlarged views show that our model better preserves vessel continuity and local topological structure than the competing methods, with fewer discontinuities and missing vessel segments, especially in thin branches and bifurcation regions.

Table 1. Quantitative comparison on MOSXAV and ICA_NJ

| *Dataset* | *Method* | *IoU* | *Dice* | *Sen.* | *Pre.* |
|---|---|---|---|---|---|
| ***MOSXAV*** | ***U-Net*** | 0.601±0.134 | 0.741±0.113 | 0.769±0.132 | <u>0.738±0.159</u> |
| | ***UNet++*** | 0.522±0.170 | 0.669±0.150 | 0.739±0.095 | 0.647±0.227 |
| | ***nnU-Net*** | 0.511±0.174 | 0.658±0.157 | 0.779±0.104 | 0.608±0.216 |
| | ***ResUNet*** | 0.587±0.142 | 0.729±0.124 | 0.798±0.141 | 0.696±0.150 |
| | ***ResNet-FPN*** | 0.597±0.115 | 0.741±0.095 | 0.781±0.115 | 0.727±0.141 |
| | ***Swin-UNet*** | 0.587±0.146 | 0.729±0.125 | 0.765±0.114 | 0.722±0.184 |
| | ***Swin-UMamba*** | <u>0.601±0.133</u> | <u>0.742±0.112</u> | **0.837±0.079** | 0.693±0.177 |
| | ***Ours*** | **0.645±0.115** | **0.778±0.092** | <u>0.815±0.121</u> | **0.767±0.135** |
| ***ICA_NJ*** | ***U-Net*** | 0.811±0.071 | 0.894±0.048 | 0.890±0.066 | 0.902±0.061 |
| | ***UNet++*** | 0.816±0.062 | 0.897±0.040 | 0.898±0.053 | 0.900±0.054 |
| | ***nnU-Net*** | 0.815±0.064 | 0.897±0.042 | 0.902±0.051 | 0.894±0.058 |
| | ***ResUNet*** | 0.822±0.069 | 0.901±0.048 | 0.894±0.061 | **0.910±0.057** |
| | ***ResNet-FPN*** | 0.813±0.066 | 0.895±0.046 | 0.889±0.059 | 0.904±0.057 |
| | ***Swin-UNet*** | 0.796±0.063 | 0.885±0.043 | 0.874±0.060 | 0.899±0.046 |
| | ***Swin-UMamba*** | <u>0.823±0.064</u> | <u>0.901±0.042</u> | **0.905±0.049** | 0.901±0.059 |
| | ***Ours*** | **0.830±0.064** | **0.906±0.043** | <u>0.905±0.054</u> | <u>0.909±0.057</u> |

Note: Segmentation metrics were computed for each individual test frame/image and are reported as mean ± standard deviation across test frames/images; bold and underlined values indicate the best and second-best results, respectively.

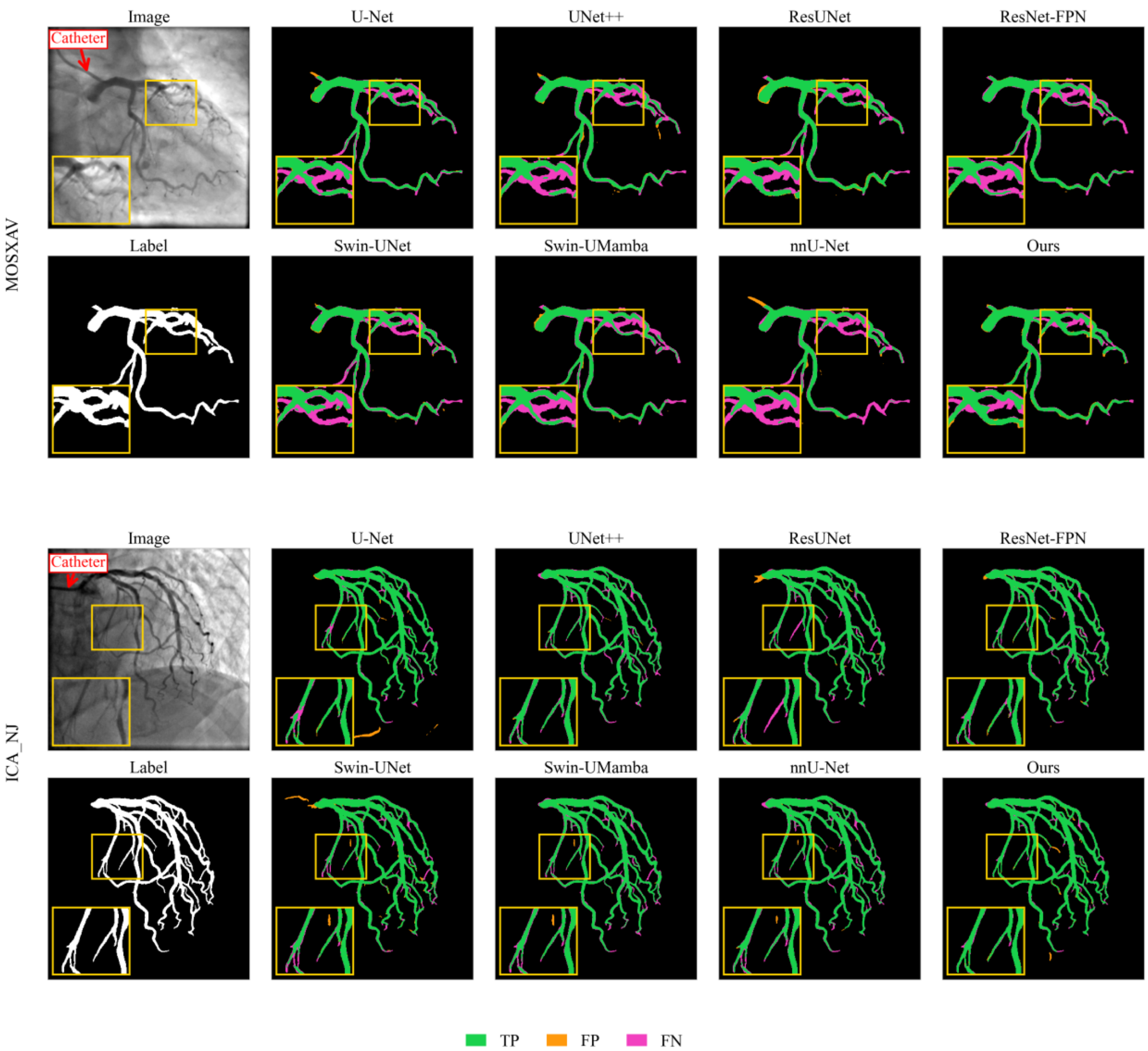

**Fig. 3 | Representative segmentation results of different methods on MOSXAV and ICA_NJ.** For each dataset, one XCA image and its reference label are shown together with predictions from eight segmentation methods. Green, orange, and magenta denote true positives (TP), false positives (FP), and false negatives (FN), respectively. The red arrows in the XCA images indicate the catheter. The yellow boxes mark connectivity-sensitive regions, which are enlarged in the lower-left corner of each panel.

### 3.3 Ablation Study and Structural Integrity Analysis

Table 2 shows that the three components are complementary: small-vessel-aware sampling primarily improves Sensitivity, EIL further improves Dice and Precision while enhancing vascular connectivity, and adding Swin yields a full model with 64.53% IoU and 77.80% Dice, representing improvements of 4.39 and 3.41 percentage points over the baseline, respectively. Single-component results are provided in Supplementary Material S2, complete topological metrics and error maps in S3, and constrained post-processing rules in S4.

Structural integrity was evaluated by comparing the ResNet-34 U-Net baseline with its EIL-augmented variant on 76 frames. Fine-branch recall and clDice quantify small-vessel recovery and centerline agreement. Connected-component and endpoint count errors measure discrepancies in fragmentation and terminal structure; a higher largest connected-component ratio indicates greater connectedness, while an endpoint ratio closer to 1 indicates closer agreement with the reference endpoint count. EIL increased fine-branch recall from 0.6478 to 0.6603 and clDice from 0.7122 to 0.7275. As shown in Table 3, connected-component count error decreased from 6.14 to 5.30 (13.7%) and endpoint count error from 12.88 to 10.86 (15.7%). The largest connected-component ratio increased from 0.8070 to 0.8300, and the endpoint ratio decreased from 2.375 to 2.208. Together, these changes indicate more complete branches and improved vessel-tree continuity (Supplementary Material S3).

Table 2. Progressive ablation results of BSC-Net

| *Samp.* | *EIL* | *Swin* | *IoU* | *Dice* | *Sen.* | *Pre.* |
|---|---|---|---|---|---|---|
| – | – | – | 0.6014±0.1185 | 0.7439±0.0980 | 0.7889±0.1237 | 0.7219±0.1333 |
| ✓ | – | – | 0.6162±0.1168 | 0.7558±0.0935 | **0.8315±0.1075** | 0.7072±0.1223 |
| ✓ | ✓ | – | <u>0.6307±0.1209</u> | <u>0.7663±0.0981</u> | 0.7843±0.1173 | <u>0.7585±0.1109</u> |
| ✓ | ✓ | ✓ | **0.6453±0.1148** | **0.7780±0.0922** | <u>0.8150±0.1212</u> | **0.7672±0.1353** |

Note: Samp. denotes small-vessel-aware sampling, EIL denotes the proposed Edge-Informed Loss, and Swin denotes the Swin Transformer module. Results are computed at the individual MOSXAV test-frame level and reported as mean ± standard deviation; bold and underlined values indicate the best and second-best results, respectively.

Table 3. Topological metrics for the Edge-Informed Loss.

| *Configuration* | *Connected component error* | *Endpoint error* | *Endpoint ratio* | *Largest connected component ratio* |
|---|---|---|---|---|
| ***ResNet-34 U-Net*** | 6.14 | 12.88 | 2.375 | 0.8070 |
| ***+ Edge-Informed Loss (EIL)*** | 5.30 | 10.86 | 2.208 | 0.8300 |

## 3.4 Cross-Dataset Generalization and Computational Efficiency

Table 4. Cross-dataset training–testing Dice (%)

| ***Training domain \ Test domain*** | ***MOSXAV*** | ***ICA_NJ*** | ***MOSXAV+ICA_NJ*** |
|---|---|---|---|
| ***MOSXAV*** | **77.80** | 75.62 | 76.44 |
| ***ICA_NJ*** | 68.50 | **89.54** | 81.48 |
| ***MOSXAV+ICA_NJ*** | 75.61 | 89.42 | **83.97** |

Cross-dataset results are shown in Table 4. Dice scores for MOSXAV→ICA_NJ and ICA_NJ→MOSXAV were 75.62% and 68.50%, respectively, demonstrating a clear domain shift between the two datasets. Joint training achieved the highest Dice of 83.97% on the mixed domain. Overall, these results demonstrate improved generalization across heterogeneous training and testing domains and indicate that exposure to heterogeneous training data improves overall adaptability.

We further compared the computational complexity of ResNet-UNet and BSC-Net in terms of parameter count and FLOPs. Restricting the Swin Transformer to the low-resolution bottleneck enables long-range contextual modeling without substantially increasing the overall computational burden. Detailed complexity statistics are provided in Supplementary Material S7.

## 4. Downstream Quantitative Analysis

Using the vessel segmentation results, we calculated two groups of quantitative parameters: morphological measures describing vessel diameter and stenosis, and image-derived hemodynamic measures describing contrast arrival, filling, and propagation along the vessel.

### 4.1 Apparent Vessel Diameter and Diameter Stenosis Ratio

The post-processed mask is skeletonized and a continuous main trunk is extracted. The EDT value at each centerline location is used to approximate the local radius, and the apparent vessel diameter is defined as:

$$D(i) = 2 \times EDT(ci) \tag{9}$$

where $c_i$ denotes the i-th sampled point on the centerline. This quantity can be interpreted as a proxy for apparent vessel diameter in a two-dimensional projection.

After applying a consistent smoothing procedure to the diameter curve, the approximate diameter stenosis ratio is calculated using the minimum diameter at the candidate lesion (MLD-like) and the mean diameter of the proximal and distal reference segments (RVD-like):

$$\%DS = \frac{D_{ref} - D_{min}}{D_{ref}} \times 100\% \tag{10}$$

A continuous diameter curve can localize focal diameter abnormalities and describe longitudinal changes, thereby converting vessel-segmentation results into morphometric measures for candidate stenosis detection and within-case morphological comparison. In this study, apparent diameter in the two-dimensional projection is used for relative analysis; future incorporation of DICOM geometric information or catheter-based scale calibration could extend the method to standardized QCA measurements.

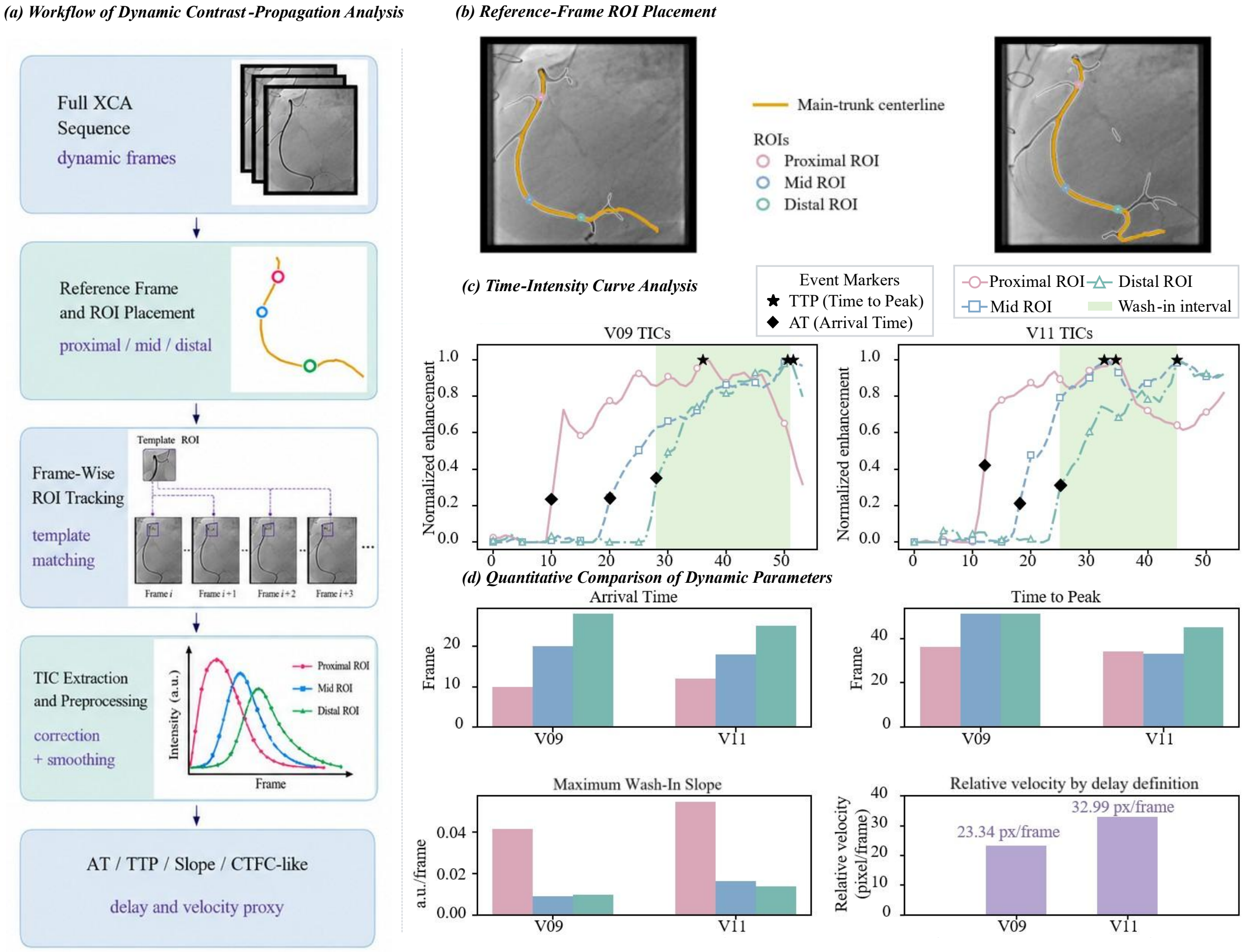

**Fig. 4 | Centerline-based apparent diameter and stenosis analysis.** (**a**) Workflow of vessel refinement, main-trunk centerline extraction, diameter profiling, and stenosis estimation. (**b**) Main-trunk centerlines and candidate stenotic sites in V09 and V11. (**c**) Centerline-based apparent diameter profiles with reference diameter, lesion interval, and MLD. (**d**) Quantitative comparison of RVD, MLD, and maximum diameter stenosis ($DS_{max}$).

As shown in Fig. 4, the segmented vessel masks are further used for main-trunk centerline extraction and centerline-based diameter analysis. Both V09 and V11 exhibit focal distal narrowing, with different reference diameters and longitudinal diameter patterns. The estimated $\boldsymbol{RVD}$, $\boldsymbol{MLD}$, and $\boldsymbol{DS_{max}}$ are 6.16 px, 2.30 px, and 62.66% for V09, and 15.15 px, 4.14 px, and 72.67% for V11, respectively. These two illustrative cases show that continuous vessel segmentation can provide a basis for centerline-based diameter analysis, candidate stenosis localization, and within-case morphological comparison.

## 4.2 Image-Derived Hemodynamic Parameter Estimation

To estimate image-derived hemodynamic parameters, proximal, middle, and distal ROIs are placed at normalized positions along the main trunk in the reference frame and tracked frame by frame using template matching. Time–intensity curves (TICs), which describe temporal changes in contrast intensity at each vascular region, are obtained after background correction, grayscale-polarity alignment, smoothing, and baseline correction. From each TIC, arrival time (AT) is defined as the first time at which the normalized curve continuously reaches 20% of its peak, reflecting contrast arrival; time to peak (TTP) is defined as the peak frame after AT, characterizing peak filling time; and the maximum wash-in slope is calculated as the largest adjacent-frame increment between AT and TTP, reflecting the rate of contrast filling. The proximal-to-distal AT difference is further defined as a CTFC-like index to characterize propagation delay, while dividing the main-trunk centerline arc length by this delay yields a relative propagation velocity in pixels/frame.

These parameters enable comparative assessment of coronary contrast filling and propagation across vascular regions, cases, or examination time points, and may support the evaluation of delayed or impaired coronary filling.

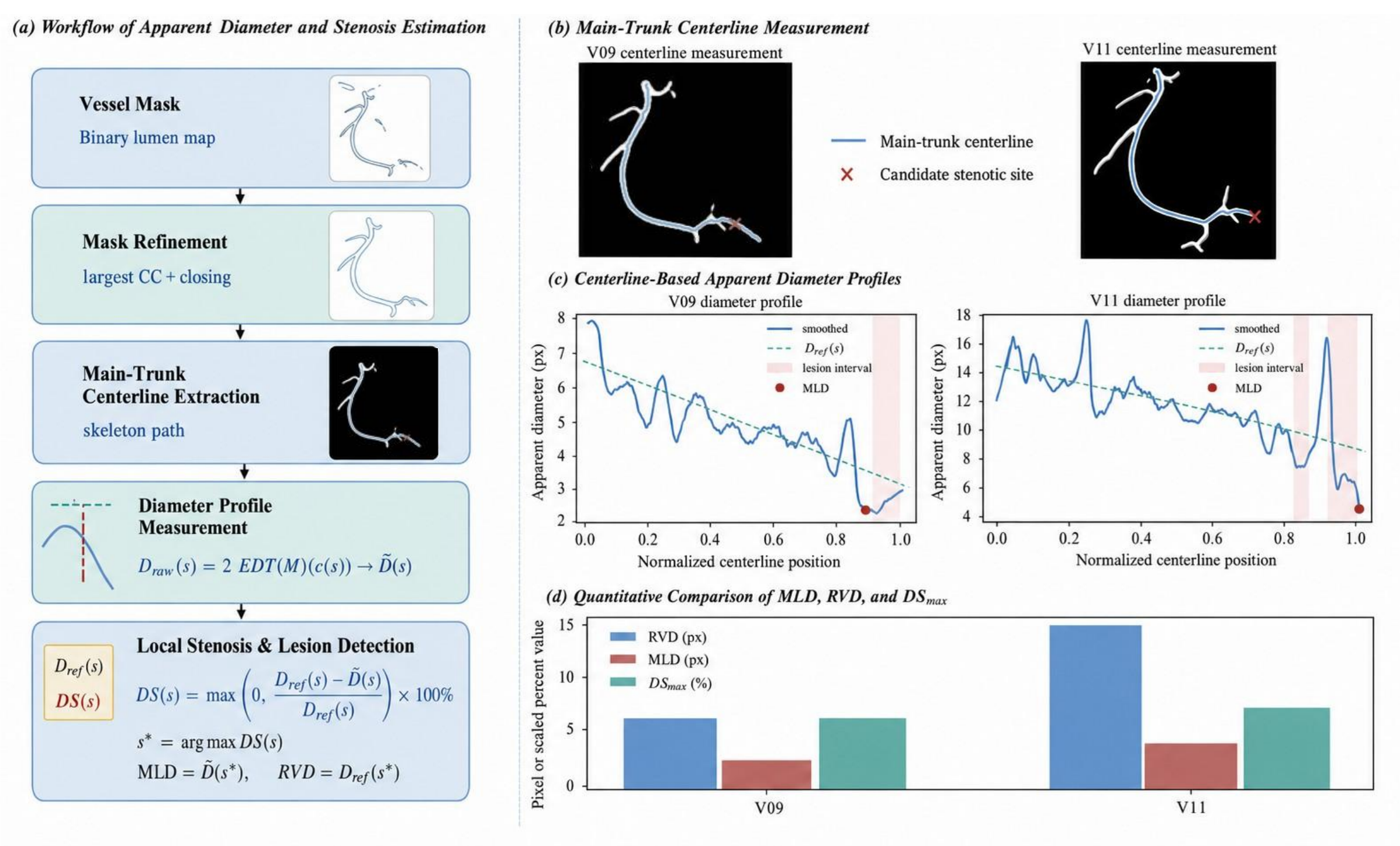

**Fig. 5. Dynamic contrast-propagation analysis in V09 and V11.** (**a**) Workflow from the full XCA sequence to reference-frame ROI placement, frame-wise ROI tracking, TIC extraction and preprocessing, and dynamic parameter calculation. (**b**) Main-trunk centerlines and proximal, middle, and distal ROIs in the reference frames. (**c**) Normalized time–intensity curves of the three ROIs with AT and TTP markers and the wash-in interval. (**d**) Quantitative comparison of AT, TTP, maximum wash-in slope, and relative velocity derived from the proximal-to-distal AT delay (CTFC-like).

In both V09 and V11, AT and TTP increased sequentially from the proximal to the middle and distal ROIs, demonstrating progressive contrast propagation along the main coronary trunk (Fig. 5). The proximal-to-distal AT delays (CTFC-like) were 18 and 13 frames for V09 and V11, respectively, while the corresponding TTP differences were 15 and 11 frames. When the propagation delay was combined with centerline path length, the estimated relative velocity was 23.34 pixels/frame for V09 and 32.99 pixels/frame for V11, with the latter being approximately 41.35% higher. The maximum wash-in slopes further reflected regional differences in the rate of contrast filling. Overall, the spatially ordered ROIs, temporally shifted TICs, and derived propagation parameters showed consistent patterns, supporting the use of segmentation-derived vessel paths for quantitative analysis of coronary contrast propagation.

## 5. Conclusion and Discussion

Accurate and structurally continuous vessel segmentation is a prerequisite for reliable quantitative analysis of coronary angiography. In this study, BSC-Net addresses this requirement through three aspects: small-vessel-aware training-sample construction, local structural supervision with EIL, and long-range context modeling at the low-resolution bottleneck. It achieves 77.80% Dice and 64.53% IoU on MOSXAV, outperforming the comparison baselines, and reaches 90.60% Dice and 83.00% IoU on ICA_NJ. Cross-domain experiments further demonstrate the generalization performance of BSC-Net across MOSXAV and ICA_NJ. Ablation experiments show the complementary contributions of the proposed components to segmentation performance. Computational complexity analysis shows that adding the Swin branch increases FLOPs by only 4.71%.

Based on the successfully delineated continuous vessel boundaries, we further conducted downstream morphometric and hemodynamic characterization, including apparent diameter/stenosis ratio, AT, TTP, CTFC-like, and relative velocity. These results show how improved vascular continuity supports downstream diameter and stenosis measurements and contrast-propagation analysis, providing a foundation for automated quantitative XCA analysis and subsequent joint anatomical–functional modeling.

This study has several limitations. The experiments are mainly based on two public datasets, and external validation across multiple centers, devices, and imaging protocols is still required. The downstream morphological and hemodynamic parameters are derived from two-dimensional images in pixel and frame units. Their agreement with clinical reference measurements, including standard QCA and TIMI frame count, and their relationship to physiological indices such as FFR require further validation. Future work will incorporate geometric calibration, more robust dynamic tracking, and larger-scale clinical data to further evaluate their reliability.

## 6. Acknowledgments

The authors gratefully acknowledge the support and assistance provided by all team members and collaborators throughout the course of this work.

## Data and Code Availability

The datasets used in this study are publicly available. The source code for BSC-Net is available at https://github.com/liwx-deeplearning/BSC-Net.

# Supplementary Information

Wanxian Li[1,#], Jiaqian Qin[1,#], Qingyi Xian[1], Yazhi Li[1], Song Chen[2], Liman Li[3], Hao He[1*]

[1]School of Biomedical Engineering, Sun Yat-sen University, Shenzhen, 518107, China.

[2]Department of Cardiovascular Surgery, Zhongnan Hospital of Wuhan University, Wuhan, 430071, China

[3]Department of Laboratory Medicine, West China Hospital of Sichuan University, Chengdu, 610041, China.

[†] Correspondence: H.H. (hehao23@mail.sysu.edu.cn)

# These authors contributed equally.

# S1. Supplementary Implementation and Evaluation Details

MOSXAV contains 30 dynamic XCA sequences with 1,789 frames, including 844 paired image/mask annotations. Vessel and balloon annotations were merged into the foreground, whereas catheter and other regions were treated as background. Device-related annotation discontinuities were corrected before sample construction. Peak-opacification frames were identified according to vessel-foreground area, and five-frame windows centered on these frames were extracted, yielding 222 static image-mask pairs (146 training and 76 test samples). ICA_NJ contains 616 paired static images; following the original 4:1 partition, 492 images formed the training pool and 124 the test set, and the training pool was further divided into 393 training and 99 validation images. All samples followed the original sequence/case-level split protocol. Training augmentation included random flipping, rotation, scaling, translation, and grayscale perturbation.

The implementation environment consisted of Python 3.10.20, PyTorch 2.6.0, and CUDA 11.8, with automatic mixed precision enabled. BSC-Net was trained on a single NVIDIA GeForce RTX 3090 for 50 epochs with a batch size of 2 using AdamW. All trainable models used the same data split, 512×512 input resolution, augmentation strategy, and validation-based model-selection protocol; the test set was not used for hyperparameter selection.

Unless otherwise specified, quantitative results are reported as mean ± standard deviation.

## S1.1 Evaluation Metrics for Vascular Topology

In addition to Dice, clDice, and main-trunk/fine-branch recall, we further use connected-component count error, endpoint count error, endpoint count ratio, and largest connected-component ratio to evaluate the structural integrity of the predicted vessel tree. Let the predicted binary vessel mask be $P$ and the ground-truth annotation be $G$.

**1. Connected Component Count Error**

Using 8-connectivity, the numbers of connected components in the predicted mask and the ground truth are first counted. Let

$$N_{\mathrm{CC}}(P), \qquad N_{\mathrm{CC}}(G) \tag{1}$$

denote the numbers of connected components in the prediction and ground truth, respectively. The connected-component count error is defined as

$$E_{\mathrm{CC}} = |N_{\mathrm{CC}}(P) - N_{\mathrm{CC}}(G)| \tag{2}$$

This metric measures the discrepancy between the predicted vessel tree and the ground truth in terms of global connectivity. A larger $E_{\mathrm{CC}}$ usually indicates more vessel breaks, fragmented predictions, or additional isolated false-positive regions; therefore, lower values are better.

**2. Endpoint Count and Endpoint Count Error**

The binary vessel mask is first skeletonized to obtain a one-pixel-wide vessel centerline

$$S(M) = \mathrm{Skeletonize}(M) \tag{3}$$

where $M$ denotes an arbitrary binary vessel mask.

For a pixel $x$ on the skeleton, let the number of other skeleton pixels within its 8-neighborhood be

$$d(x) = \sum_{y \in \mathcal{N}_8(x)} S\,(M)(y) \tag{4}$$

A skeleton pixel is defined as an endpoint when it has exactly one neighboring skeleton pixel, i.e.,

$$x \in E(M) \Leftrightarrow S(M)(x) = 1 \wedge d(x) = 1 \tag{5}$$

Accordingly, the number of endpoints in mask $M$ is

$$N_{\mathrm{EP}}(M) = \sum_x \mathbf{1}\,[S(M)(x) = 1 \ \wedge\ d(x) = 1] \tag{6}$$

The endpoint count error of the prediction relative to the ground truth is defined as

$$E_{\mathrm{EP}} = |N_{\mathrm{EP}}(P) - N_{\mathrm{EP}}(G)| \tag{7}$$

This metric reflects the discrepancy in branch-termination structure between the predicted vessel tree and the ground truth. Vessel breaks usually introduce additional endpoints, whereas missed small branches may also alter the endpoint count. Therefore, a smaller $E_{\mathrm{EP}}$ indicates that the predicted topology is closer to the ground-truth vessel tree.

**3. Endpoint Count Ratio**

To further evaluate how well the number of endpoints is preserved relative to the ground truth, the endpoint count ratio is defined as

$$R_{\mathrm{EP}} = \frac{N_{\mathrm{EP}}(P)}{N_{\mathrm{EP}}(G)} \tag{8}$$

Ideally,

$$R_{\mathrm{EP}} \approx 1 \tag{9}$$

indicating that the prediction and ground truth have similar numbers of endpoints.

When

$$R_{\mathrm{EP}} > 1 \tag{10}$$

the prediction generally contains more endpoints, which may be associated with local vessel breaks or fragmentation;

when

$$R_{\mathrm{EP}} < 1 \tag{11}$$

the result may instead indicate missed terminal branches or merging of branch structures.

Following the implementation used in this study, when the ground truth contains no endpoints, i.e., $N_{\mathrm{EP}}(G) = 0$, the following special definition is adopted:

$$R_{\mathrm{EP}} = \begin{cases} 1, & N_{\mathrm{EP}}(P) = 0, \\ N_{\mathrm{EP}}(P), & N_{\mathrm{EP}}(P) > 0, \end{cases} \quad \text{when } N_{\mathrm{EP}}(G) = 0 \tag{12}$$

Thus, the optimal value of this metric is neither the largest nor the smallest possible value, but rather the value closest to 1.

**4. Largest Connected Component Ratio (LCC Ratio)**

For the predicted vessel mask $P$, suppose that 8-connected-component analysis yields $K$ connected components

$$C_1, C_2, \dots, C_K \tag{13}$$

where $|C_k|$ denotes the number of foreground pixels in the $k$-th connected component. The total predicted foreground area is

$$A_P = \sum_{k=1}^{K} |C_k| \tag{14}$$

The area of the largest connected component is

$$A_{\max} = \max_{1 \le k \le K} |C_k| \tag{15}$$

The largest connected component ratio is defined as

$$R_{\mathrm{LCC}} = \frac{A_{\max}}{A_P} \tag{16}$$

When the prediction contains no foreground pixels, we define

$$R_{\mathrm{LCC}} = 0 \tag{17}$$

$R_{\mathrm{LCC}}$ measures the extent to which predicted vessel pixels are concentrated in a single dominant connected structure, with the range $0 \le R_{\mathrm{LCC}} \le 1$.

A higher $R_{\mathrm{LCC}}$ indicates that a larger proportion of predicted vessel pixels belongs to the same dominant connected tree and that fragmentation is lower; therefore, higher values are generally better.

**5. Dataset-Level Statistics**

The topology metrics above are first computed independently for each test image and then aggregated over all test frames. For any frame-wise metric $m_i$, the dataset-level mean is

$$\bar{m} = \frac{1}{N}\sum_{i=1}^{N} m_i \tag{18}$$

where $N$ is the number of test frames included in the evaluation.

# S2. Single-Component Ablation Study

Table S1. Ablation results when each component is independently added to the baseline.

| *Samp.* | *EIL* | *Swin* | *IoU* | *Dice* | *Sen.* | *Pre.* |
|---|---|---|---|---|---|---|
| – | – | – | 0.6014±0.1185 | 0.7439±0.0980 | <u>0.7889±0.1237</u> | 0.7219±0.1333 |
| ✓ | – | – | <u>0.6162±0.1168</u> | <u>0.7558±0.0935</u> | **0.8315±0.1075** | 0.7072±0.1223 |
| – | ✓ | – | **0.6210±0.1260** | **0.7584±0.1004** | 0.7711±0.1195 | **0.7618±0.1276** |
| – | – | ✓ | 0.6117±0.1359 | 0.7498±0.1109 | 0.7875±0.1060 | <u>0.7396±0.1672</u> |

Note: Each component is independently added to the same ImageNet-pretrained ResNet-34 U-Net baseline. Samp. denotes small-vessel-aware sampling, EIL denotes the Edge-Informed Loss, and Swin denotes the Swin Transformer module. Results are reported as mean ± standard deviation.

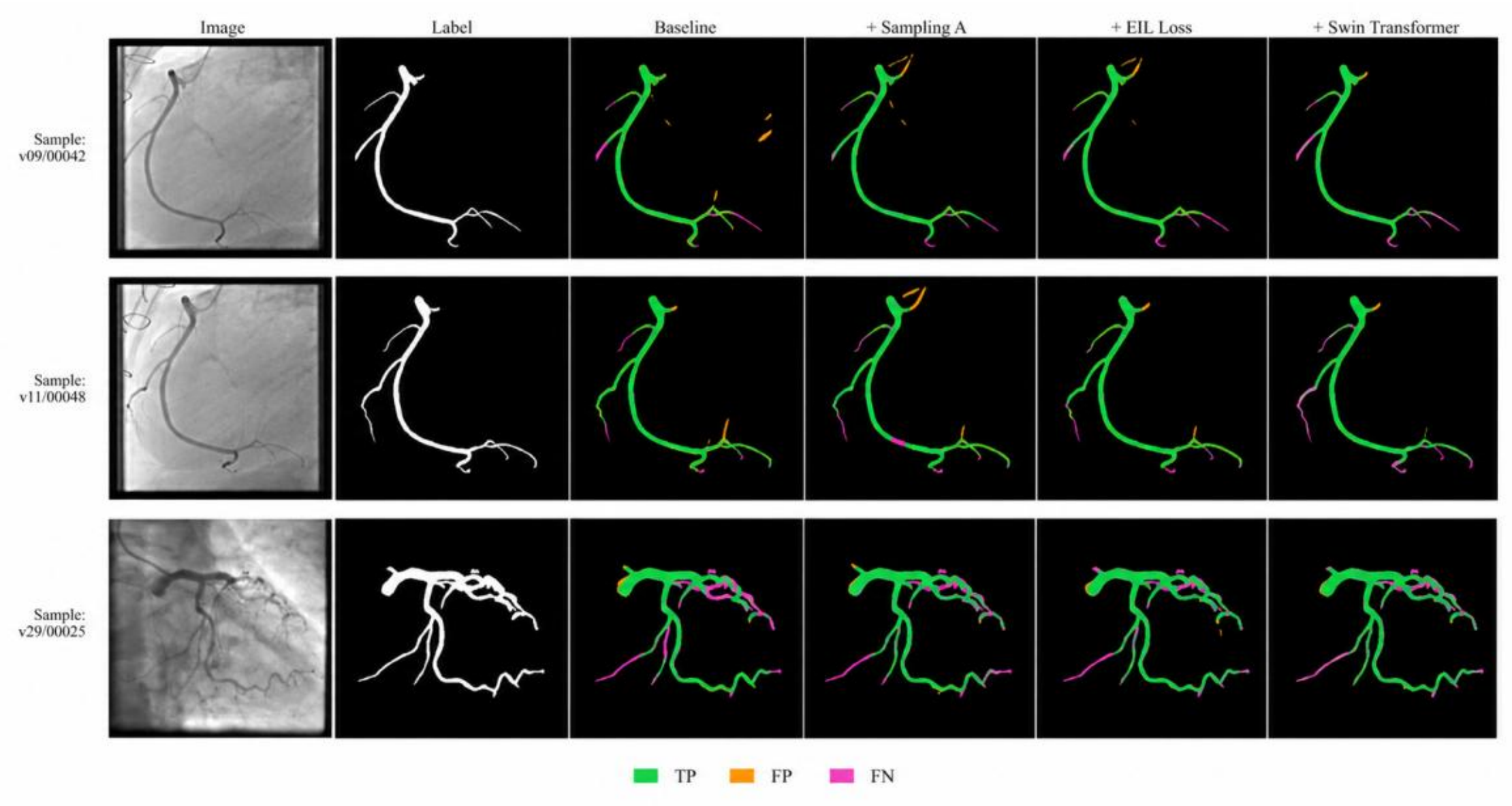

Fig. S1 | Representative visualizations of the single-component ablation study.

The independent ablation results show that small-vessel-aware sampling primarily improves distal fine-branch recall, although it may introduce a small number of additional false positives. EIL is more effective in reducing scattered false detections and local discontinuities, whereas the Swin Transformer provides additional improvement at complex bifurcations and weakly opacified regions. The progressive combination of the three components is reported in Table 2 of the main manuscript and is therefore not repeated here.

# S3. Additional Topological Visualization and Structural Analysis

Across the 76 frames used for structural evaluation, EIL increased fine-branch recall from 0.6478 to 0.6603 and clDice from 0.7122 to 0.7275. Fig. S2 provides additional quantitative visualization and representative topology-sensitive examples complementing Table 3 of the main manuscript.

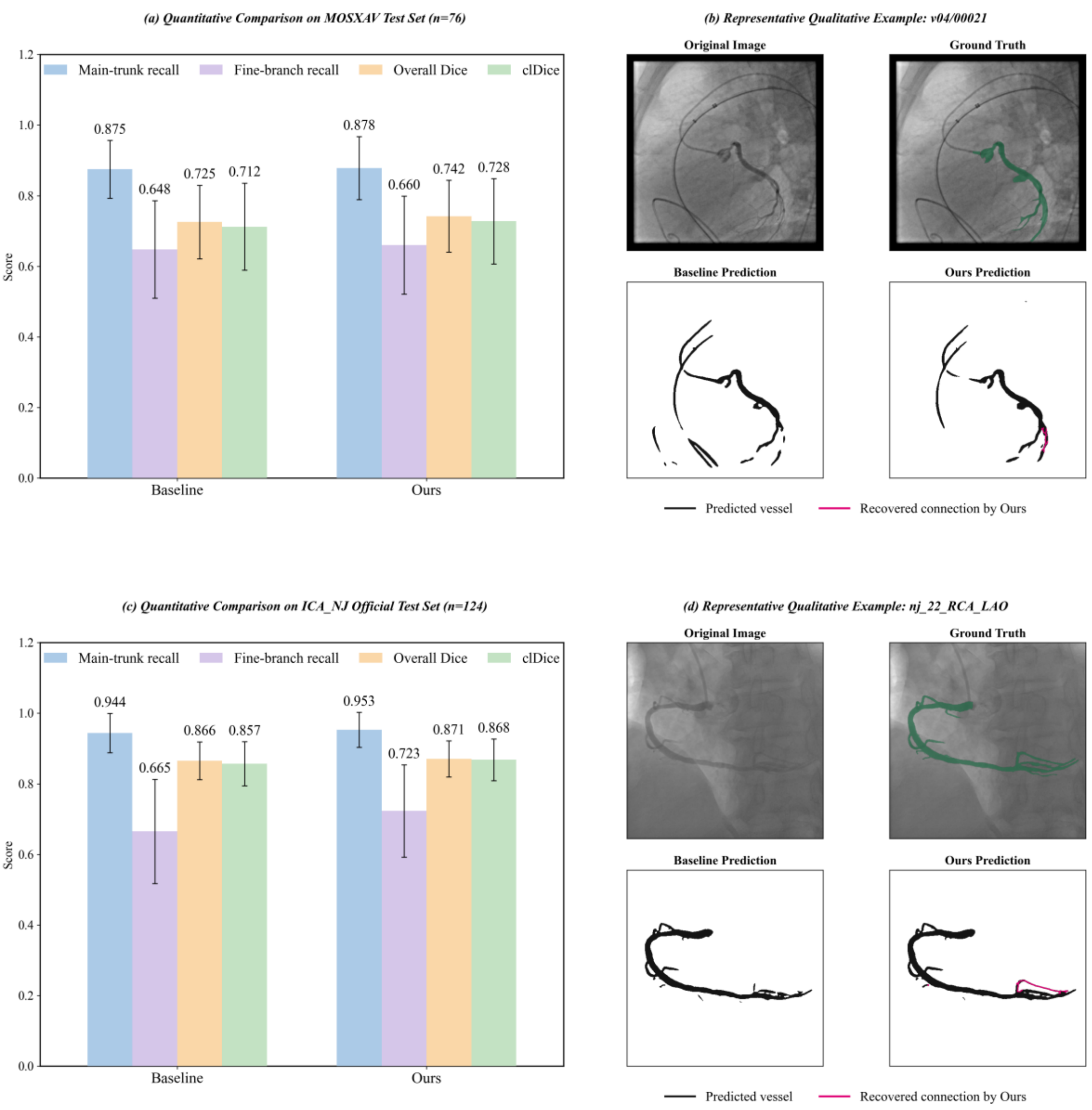

Fig. S2 | Comparison of topology metrics and error maps for representative cases from MOSXAV and ICA_NJ.

While largely preserving main-trunk recall, EIL improves fine-branch recall and centerline consistency and reduces redundant endpoints and fragmented connected components. These results further indicate that its benefit is reflected not only in region-overlap metrics but also in the structural integrity of the overall vessel tree.

## S4. Structural-Continuity Post-processing

Distal small vessels and poorly opacified regions in X-ray coronary angiography often exhibit low contrast, making segmentation masks susceptible to local omissions or breaks between adjacent vessel segments. Such discontinuities can cause premature centerline termination and affect vessel-path extraction, region-of-interest placement, and subsequent downstream quantitative analysis. Therefore, after BSC-Net inference, we apply a structural-continuity post-processing procedure that performs limited reconnection only for local short-range breaks supported by probabilistic evidence.

As shown in Fig. S3, the procedure takes the network probability map and binary mask as inputs and sequentially performs mask normalization, topology extraction, endpoint-geometry estimation, candidate-path search, candidate validation, iterative updating, and width reconstruction, ultimately producing a continuous vessel mask for downstream analysis.

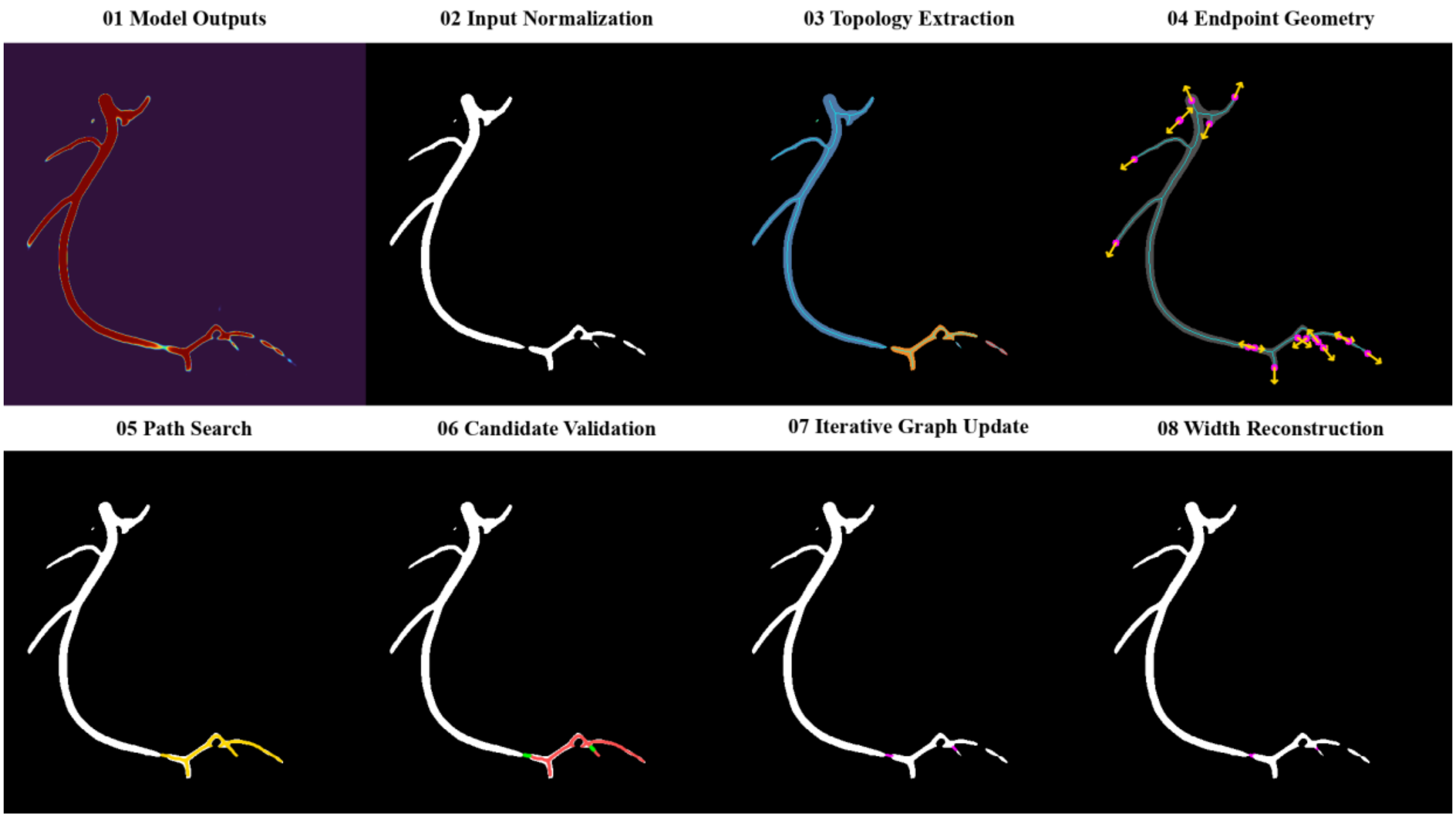

Fig. S3 | Structural-continuity post-processing pipeline for the segmentation mask. After binarization of the network output, the skeleton, connected components, and endpoints are extracted. Candidate reconnection paths are then searched and filtered using the probability map. The topology is updated after each accepted connection, and path width is reconstructed according to the local vessel scale.

Let the pixel-wise vessel probability map output by the network be $P$ and the initial thresholded binary mask be $B_0$. The goal of post-processing is to repair short-range discontinuities using weak vessel responses retained in the probability map, without substantially changing the foreground area, so that the repaired mask is more suitable for centerline extraction and downstream analysis.

### S4.1 Skeletonization, Connected Components, and Endpoint Detection

The current binary mask is first skeletonized to one-pixel width, and vessel components are labeled using 8-connectivity. For an arbitrary skeleton pixel $p$, the skeleton degree and candidate endpoint set are defined as

$$d_S(p) = \sum_{q \in \mathcal{N}_8(p)} S(q), \qquad E = \{p \in \Omega \mid S(p) = 1,\ d_S(p) = 1\} \tag{19}$$

where $S$ is the one-pixel-wide vessel skeleton, $\Omega$ is the image domain, $\mathcal{N}_8(p)$ denotes the 8-neighborhood of pixel $p$, $d_S(p)$ is the skeleton degree, and $E$ is the candidate endpoint set. A Euclidean distance transform is also applied to the mask to estimate the local radius from each skeleton point to the vessel boundary.

## S4.2 Endpoint Extension-Direction Estimation

For each candidate endpoint $e \in E$, a fixed-length trace is followed inward along its connected skeleton component to obtain a local reference point $q_e$. The unit outward extension vector of the endpoint is

$$\mathbf{v}_e = \frac{e - q_e}{\|e - q_e\|_2} \tag{20}$$

where $e$ is the current endpoint, $q_e$ is the reference point obtained by tracing inward along the skeleton, and $\|\cdot\|_2$ denotes the Euclidean norm. Candidate targets are preferentially selected within the forward search region indicated by $\mathbf{v}_e$ and are required to belong to a different connected component, thereby reducing the risk of short-circuit self-connections within the same component.

## S4.3 Probability-Guided Candidate-Path Search

An 8-neighbor minimum-cumulative-cost path search is performed between an endpoint and each candidate target. The traversal cost at pixel $x$ is defined as

$$C(x) = -\ln\hat{P}(x) + \lambda_p\, \mathbf{1}[P(x) < \tau_s], \qquad \hat{P}(x) = \max(P(x), 10^{-5}) \tag{21}$$

where $P(x)$ is the vessel probability at pixel $x$, $\hat{P}(x)$ is the clipped probability used to avoid logarithmic overflow, $\mathbf{1}[\cdot]$ is the indicator function, $\tau_s = 0.20$ is the probability-support threshold, and $\lambda_p = 25$ is the additional penalty for low-probability regions. Consequently, the search preferentially traverses regions that retain vessel responses, whereas paths crossing background regions incur higher costs.

## S4.4 Candidate-Connection Scoring

For a searched candidate path $\Gamma$, paths that cross unrelated components, exhibit obvious reversal, or lack sufficient probability support are first discarded. The remaining paths are scored as follows:

$$s(\Gamma) = \mu_\Gamma + 0.25\rho_\Gamma - 0.004|\Gamma| + 0.05\, \mathbf{1}[A^{c_s} < A^{c_t}] \tag{22}$$

where $\mu_\Gamma$ is the mean vessel probability over the portion of the candidate path missing from the original mask, $\rho_\Gamma$ is the proportion of path pixels satisfying $P(x) \geq \tau_s$, $|\Gamma|$ is the path length, and $A^{c_s}$ and $A^{c_t}$ are the areas of the source and target components, respectively. The score favors short paths with high probabilities and a high support ratio, while giving slight priority to connections from smaller vessel components to larger components.

## S4.5 Candidate-Acceptance Constraints and Iterative Updating

Candidate paths are attempted in descending order of $s(\Gamma)$. Adding a path to the current mask $B$ yields a temporary mask $B'$; the connection is accepted only when both of the following conditions are satisfied:

$$\frac{|B'| - |B_0|}{|B_0|} \leq \eta_A, \qquad K_8(B') < K_8(B) \tag{23}$$

where $|B|$ denotes the number of foreground pixels in a mask, $B_0$ is the initial mask before post-processing, $\eta_A = 0.08$ is the maximum cumulative foreground-area growth ratio, and $K_8(\cdot)$ denotes the number of 8-connected components. The first constraint limits the cumulative foreground-area increase to within $\eta_A$, whereas the second ensures that the accepted path actually improves topological connectivity. After each accepted path, the skeleton, endpoints, and connected components are recomputed before the next search iteration.

### S4.6 Local Vessel-Width Reconstruction

The minimum-cost path specifies only the center trajectory of a connection. Let $\Gamma = \{x_0, \dots, x_{L-1}\}$ be an accepted path, and let $r_s$ and $r_t$ denote the local radii at the two connection ends estimated by the distance transform. The reconstruction radius at the $i$-th pixel along the path is defined as

$$r_i = (1 - t_i)r_s + t_i r_t, \qquad t_i = \frac{i}{L-1}, \qquad i = 0, \dots, L-1 \tag{24}$$

Thus, the connection width varies linearly between the vessel scales at the two ends. For the local disk centered at $x_i$ with radius $r_i$, only pixels satisfying $P(x) \geq \tau_s$ are added to the mask, thereby avoiding pronounced thickening in low-probability background regions. Post-processing terminates when no acceptable candidate is produced in the current iteration or when the preset maximum number of iterations is reached.

This post-processing procedure is used only to prepare masks for downstream quantitative analysis and does not participate in BSC-Net training. Its purpose is to repair short-range vessel discontinuities caused by local low contrast when they remain supported by probabilistic evidence, thereby reducing premature centerline termination and providing vessel masks with improved anatomical continuity for subsequent downstream quantitative analysis.

### S4.7 Representative Repair Results

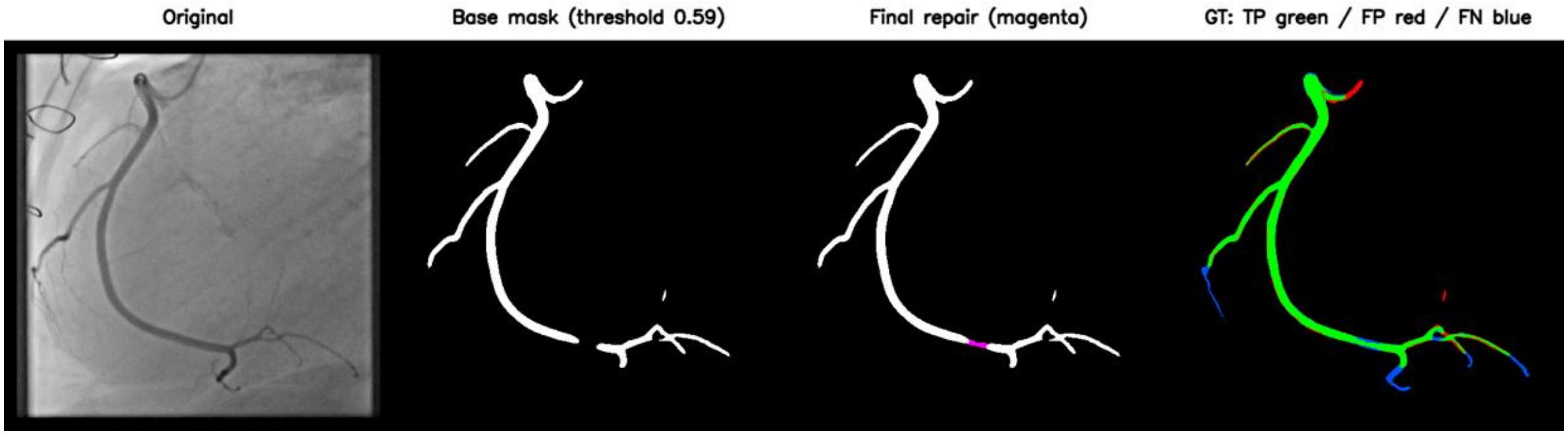

Fig. S4 | Example of local discontinuity repair by post-processing.

The representative example shows fewer local false negatives after repair without an obvious large-area increase in false positives. Low-contrast terminal segments lacking sufficient probability support remain unconnected, reducing the risk of erroneous reconnection.

## S5. Detailed Computation of Edge-Informed Loss

The main manuscript retains only the core definition and key equations of EIL. For reproducibility, the complete computation of vessel-neighborhood construction, directional gradient residuals, radial frequency weighting, and neighborhood normalization is provided below.

For implementation, max pooling with a $7 \times 7$ kernel, stride 1, and padding 3 is applied to the ground-truth vessel label $Y$ to obtain the vessel-neighborhood mask $R$.

For direction $a \in \{x, y\}$, the vessel neighborhood $R$ is shifted by one pixel in the corresponding direction and multiplied by the original neighborhood to obtain the directional gate $G_a$. The directional gradient residual between the predicted probability map $P$ and ground-truth label $Y$ is then computed within the gated region:

$$G_a = R \odot T_a(R), \qquad E_a = G_a \odot (\nabla_a P - \nabla_a Y), \qquad a \in \{x, y\} \tag{25}$$

where $T_a(\cdot)$ denotes a one-pixel shift along direction $a$, $\nabla_a$ denotes the corresponding forward difference, $\odot$ denotes element-wise multiplication, and $G_a$ and $E_a$ are the directional gate and gradient residual, respectively.

For the $b$-th sample and direction $a$, an orthonormally normalized two-dimensional real fast Fourier transform is applied to $E_{a,b}$, and the radial frequency weight is computed as

$$\hat{E}_{a,b}(u,v) = \mathrm{RFFT}_2^{\mathrm{ortho}}(E_{a,b})(u,v), \qquad W(u,v) = \left[f_x(v)^2 + f_y(u)^2 + \varepsilon_f\right]^{\kappa/2} \tag{26}$$

where $(u,v)$ denotes a discrete frequency location, $f_x(v)$ and $f_y(u)$ are the discrete horizontal and vertical frequencies, respectively, $\varepsilon_f$ is a numerical-stability constant, and $\kappa = 1$ is the radial frequency-weighting exponent used in this study.

For the $b$-th sample, the frequency-weighted gradient-residual energy in direction $a$ is defined as

$$\mathcal{E}_{a,b} = \frac{1}{|\Omega_f|}\sum_{(u,v)\in\Omega_f} W(u,v)\left|\hat{E}_{a,b}(u,v)\right|^2 \tag{27}$$

where $\Omega_f$ denotes the set of discrete frequency locations in the two-dimensional real spectrum and $|\Omega_f|$ is the number of frequency locations in this set.

Finally, the horizontal and vertical frequency-domain energies are summed and normalized by the vessel-neighborhood occupancy ratio $q_b$ for each sample:

$$q_b = \frac{1}{HW}\sum_{i=1}^{H}\sum_{j=1}^{W} R_{b,i,j}\,, \qquad \mathcal{L}_{\mathrm{EIL}} = \frac{1}{B}\sum_{b=1}^{B}\frac{\mathcal{E}_{x,b}+\mathcal{E}_{y,b}}{\max(q_b,\varepsilon)} \tag{28}$$

# S6. Implementation Details of the Swin Transformer Bottleneck

The main manuscript reports only the bottleneck configuration directly relevant to BSC-Net. Additional implementation details of W-MSA/SW-MSA, shifted windows, and residual fusion are provided below for complete reproducibility.

The bottleneck feature $C_5$ output by the ResNet-34 encoder has a size of $B \times 512 \times 16 \times 16$. A $1 \times 1$ convolution first maps the channel dimension from 512 to 384, after which the $16 \times 16$ spatial dimensions are flattened into 256 tokens, yielding:

$$T_0 \in \mathbb{R}^{B\times 256\times 384} \tag{29}$$

As shown on the left side of Fig. 2(b), each Swin Block consists of Layer Normalization, W-MSA or SW-MSA, an MLP, and two residual connections. Let $T_{l-1}$ be the input to the $l$-th Block; the feature update is expressed as

$$\tilde{T}_l = T_{l-1} + A_l(\mathrm{LN}(T_{l-1})), \qquad T_l = \tilde{T}_l + \mathrm{MLP}\left(\mathrm{LN}(\tilde{T}_l)\right) \tag{30}$$

where $\mathrm{LN}(\cdot)$ and $\mathrm{MLP}(\cdot)$ denote layer normalization and the multilayer perceptron, respectively, and $A_l(\cdot)$ denotes the window-attention operation used by the current Block.

As shown on the right side of Fig. 2(b), the bottleneck feature has a spatial size of $16 \times 16$ and the window size is set to $M = 4$. The feature map is therefore partitioned into 16 non-overlapping $4 \times 4$ windows, each containing 16 tokens. Features within each window are first linearly projected to queries $Q$, keys $K$, and values $V$; window attention can then be written uniformly as

$$\mathrm{Attn}_M(Q,K,V) = \mathrm{Softmax}\left(\frac{QK^{\top}}{\sqrt{d}} + B_{\mathrm{rel}} + M\right)V \tag{31}$$

where $d$ denotes the feature dimension of a single attention head, $B_{\mathrm{rel}}$ is the relative position bias within the window, and $M$ is the attention mask. W-MSA computes self-attention directly within fixed windows, establishing interactions among vessel-orientation, edge, and branch features within the same window while avoiding the larger computational cost of self-attention over the entire feature map.

Because W-MSA restricts feature interaction to fixed windows, vessel features located in different windows cannot directly exchange information within the same Block. To establish interactions between neighboring windows, SW-MSA cyclically shifts the two-dimensional feature map $X$ by half a window in both the horizontal and vertical directions before window partitioning. The shifting and spatial restoration after attention are expressed as:

$$\tilde{X} = \mathrm{Roll}(X; -s, -s), \qquad X_{\mathrm{out}} = \mathrm{Roll}(\tilde{X}_{\mathrm{attn}}; s, s), \qquad s = \frac{M}{2} = 2 \tag{32}$$

where $\mathrm{Roll}(\cdot)$ denotes cyclic shifting along the height and width dimensions of the feature map, $\tilde{X}$ is the shifted feature map, $\tilde{X}_{\mathrm{attn}}$ is the feature after attention within shifted windows, and $X_{\mathrm{out}}$ is the output restored to its original spatial arrangement by reverse cyclic shifting.

After shifting, the model repartitions the feature map into $4 \times 4$ windows at the new spatial positions, allowing tokens that originally lay on opposite sides of neighboring window boundaries to enter the same window and participate in self-attention. Because cyclic shifting wraps content from one feature-map boundary to the opposite side, SW-MSA adds the mask $M$ to the attention scores to prevent erroneous interactions between boundary tokens that were not spatially adjacent in the original feature map. After window attention, reverse cyclic shifting restores the original spatial arrangement.

We sequentially stack four Swin Blocks in the order W-MSA, SW-MSA, W-MSA, and SW-MSA, with window shifts of 0, 2, 0, and 2, respectively. W-MSA performs local feature interaction within fixed windows, whereas SW-MSA establishes information propagation between neighboring windows through cyclic shifting, window repartitioning, masked attention, and reverse shifting. Alternating the two attention mechanisms allows local features to propagate progressively across the original window boundaries, thereby expanding the effective interaction range of the bottleneck representation. Because each bottleneck token corresponds to a relatively large region of the input image, such cross-window propagation can integrate relationships among spatially separated coronary segments, main-trunk trajectories, and branch connections, providing broader structural context for maintaining vessel-tree morphology and continuity.

The outputs of the four Swin Blocks are restored to two-dimensional feature maps and mapped back to 512 channels using a $1 \times 1$ convolution, yielding the Swin-Transformer branch output $\Delta$. This output is then fused with the original bottleneck feature $C_5$ through a residual connection:

$$\Delta = \mathrm{Conv}_{1\times1}\left[\mathrm{Reshape}\left(\mathrm{Swin}(T_0)\right)\right], \qquad C_5{}' = C_5 + \gamma_s \Delta, \qquad \gamma_s^{(0)} = 0.1 \tag{33}$$

where $\gamma_s$ is a learnable residual scaling coefficient and $\gamma_s^{(0)}$ denotes its initial value. The enhanced bottleneck feature $C_5{}'$ is then passed to the decoder. The Swin Transformer is deployed only at the low-resolution bottleneck and restricts self-attention to local windows, limiting additional computational and parameter overhead.

## S7. Model Complexity and Computational Cost

Table S2. Comparison of model complexity.

| *Model* | *Parameters* | *FLOPs / image* |
|---|---|---|
| ResNet-34 + U-Net | 24,521,697 | 81.781 GFLOPs |
| **BSC-Net** | **32,015,610** | **85.631 GFLOPs** |

Note: Relative to the ResNet-34 U-Net baseline, BSC-Net increases the number of parameters by 30.56% and FLOPs per image by 4.71%. Because attention is applied only to the 16×16 low-resolution bottleneck feature, the increase in FLOPs is substantially smaller than the increase in parameter count.